\documentclass[letterpaper]{article} 
\usepackage[preprint]{aaai2027}  
\usepackage[hyphens]{url}  
\usepackage{graphicx} 
\usepackage{natbib}  
\usepackage{caption} 
\usepackage{algorithm}
\usepackage{algorithmic}
\usepackage{booktabs}
\usepackage{multirow}
\usepackage[table]{xcolor}
\usepackage{booktabs}
\usepackage{multirow}
\usepackage{amssymb}
\usepackage{amsmath}
\usepackage{newfloat}
\usepackage{listings}
\DeclareCaptionStyle{ruled}{labelfont=normalfont,labelsep=colon,strut=off} 
\floatstyle{ruled}
\newfloat{listing}{tb}{lst}{}
\floatname{listing}{Listing}

\usepackage{xcolor}

\title{VR3D: View-Robust 3D Representation Learning for Aerial-Ground Person Re-Identification}

\author{
    Chao Ji,
    Shiyu Xuan,
    Zechao Li\corresponding
}

\affiliations{
    School of Computer Science and Engineering, 
    Nanjing University of Science and Technology\\
    jc0612@njust.edu.cn,
    shiyu\_xuan@njust.edu.cn,
    zechao.li@njust.edu.cn
}

\begin{document}

\maketitle

\begin{abstract}

Aerial-ground person re-identification is a challenging task due to cross-platform viewpoint variations, which cause severe occlusion and geometric deformation. Existing methods attempt to learn view-invariant representations exclusively within the 2D image space, where drastic viewpoint variations cause the learned features to remain coupled with viewpoint bias. To address this, we propose VR3D, a View-Robust 3D Representation Learning framework that maps images into a unified 3D coordinate space to achieve view-independent feature interaction. Specifically, we introduce View-Robust 3D Representation Interaction, which leverages 3D priors extracted from single 2D observations to lift 2D appearance features into a canonical 3D space. VR3I employs 3D Geometry-Semantic Attention to establish interactions between 2D patches and 3D voxels from corresponding body parts based on their 3D spatial locations, effectively grounding 2D semantics within a 3D framework. In addition, as the reliability of these representations varies across samples due to viewpoint changes and 3D reconstruction errors, we introduce a Reliability-Aware Fusion that estimates sample-specific reliability and adaptively aggregates the multi-source representations. Extensive experiments on three benchmark datasets (CARGO, AG-ReID.v1, and AG-ReID.v2) demonstrate that VR3D outperforms recent methods, \emph{e.g.}, it achieves an improvement of 5.63\% in Rank-1 on CARGO. Our code will be released.

\end{abstract}

\section{Introduction}

Person re-identification aims to retrieve target persons with the same identity as a query image across different cameras and plays an important role in intelligent surveillance, public safety, and other fields~\cite{he2021transreid,li2023clip,he2024instruct,yuan2025poses,zhou2026hierarchical}. In recent years, with the widespread use of drones, aerial-ground person re-identification has attracted increasing attention. As a challenging cross-view retrieval scenario~\cite{zhang2023ground,nguyen2023aerial,nguyen2024ag,nguyen2025ag}, its central problem is learning stable and discriminative identity representations under viewpoint variations. Large viewpoint differences between aerial and ground cameras cause severe occlusion and geometric deformation, making the same person exhibit substantially different visible appearances.

\begin{figure}[!t]
\centering
\includegraphics[width=\columnwidth]{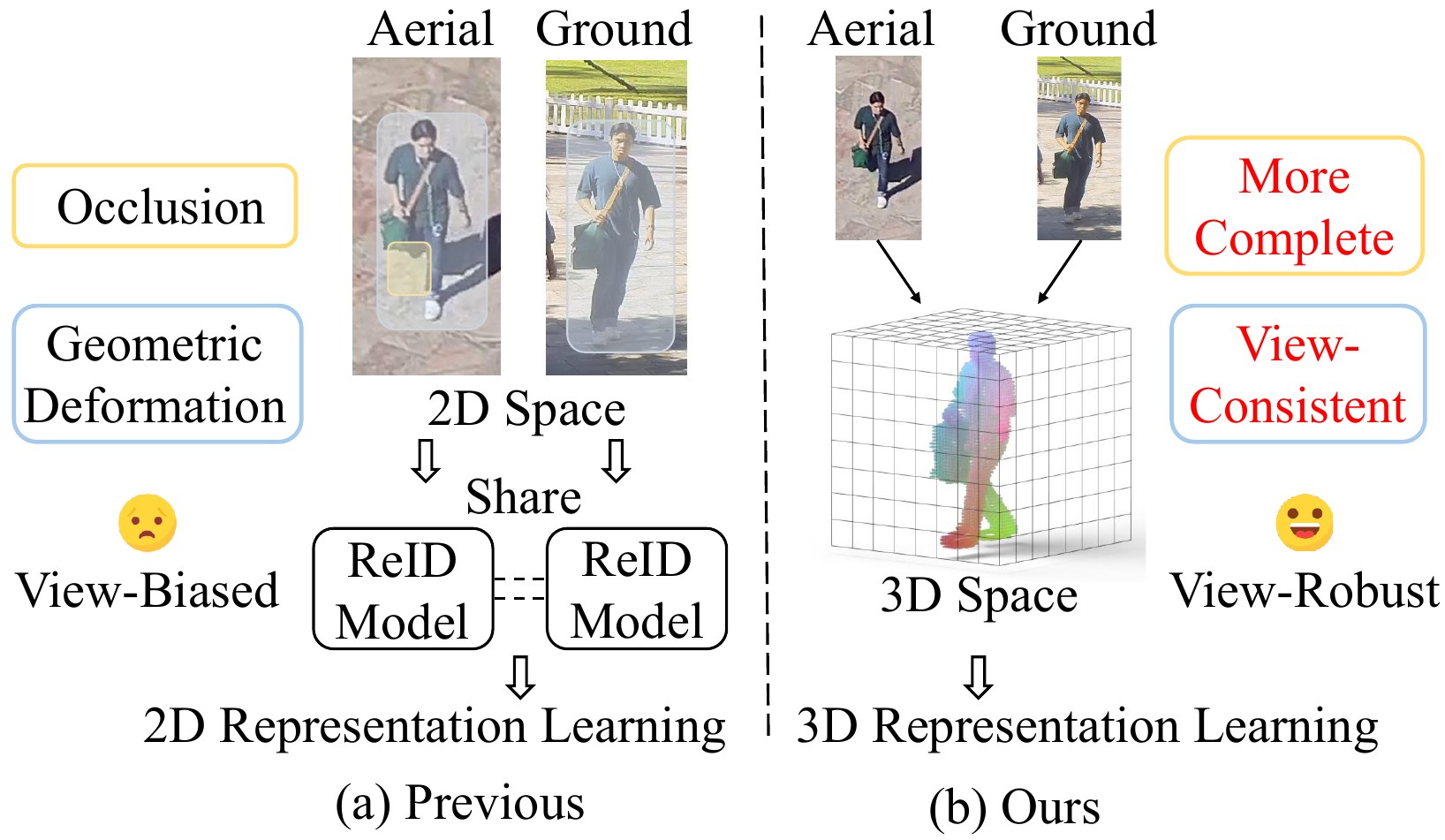}
\caption{Illustration of our motivation. (a) Previous methods learn identity representations in 2D space, where occlusion and geometric deformation introduce strong viewpoint bias. (b) Our method learns more complete and view-consistent human representations in 3D space, enabling view-robust identity learning.}
\label{fig:1}
\end{figure}

To tackle these challenges, existing methods mainly follow two paradigms: viewpoint disentanglement~\cite{zhang2024view,wang2025secap,zhang2026view} and cross-view information completion~\cite{zhang2025latex,zheng2026semantic,wang2026sd}. Although these methods achieve promising results, they attempt to learn view-invariant identity representations exclusively within the 2D image space. As shown in Fig.~\ref{fig:1} (a), a 2D image records only the visible appearance under the current viewpoint, and its representation is coupled with viewpoint bias. Therefore, it remains difficult for the model to decouple these factors and learn view-invariant identity features.

To address the difficulty of learning view-invariant features in 2D, we propose to directly achieve feature interaction and representation in a view-independent space. Visual representations in 3D space are intrinsically more robust to viewpoint changes \cite{shang2022learning}, providing a unified coordinate system that is not affected by camera angles. As illustrated in Fig.~\ref{fig:1} (b), by lifting the representation into 3D space, we mitigate the bottleneck of 2D viewpoint bias, providing more complete and consistent information to facilitate the learning of view-robust features.

Based on this insight, we propose VR3D, a View-Robust 3D Representation Learning framework designed to achieve feature interaction and representation within a unified 3D space. To establish this 3D foundation from single 2D observations, we construct an offline pipeline that leverages the pre-trained knowledge of SAM3~\cite{carion2026sam} and SAM3D~\cite{Chen_2026_CVPR} to extract 3D coordinates, camera pose parameters, and voxel representations. 

The 3D priors extracted by SAM3D capture geometric structures but lack the appearance semantics required for identity recognition. To address this, we propose the View-Robust 3D Representation Interaction (VR3I). By utilizing the extracted 3D coordinates to lift local 2D appearance features into a canonical 3D space, VR3I employs 3D Geometry-Semantic Attention to establish interactions between 2D patches and 3D voxels from corresponding body parts based on their 3D spatial locations. This grounds the 2D semantics within a 3D structural framework. Consequently, VR3I yields three distinct representations: an appearance feature extracted by a 2D image encoder, a view-robust 3D structural feature extracted by a 3D voxel encoder, and a spatially grounded feature that aligns appearance details based on 3D geometry.

However, the reliability of these three representations varies across samples. Extreme viewing angles reduce the reliability of 2D representations, while low-quality 3D reconstructions reduce the reliability of 3D-related representations. To prevent unreliable information from compromising identity matching, we further introduce a Reliability-Aware Fusion (RAF) that adaptively aggregates the three features into the final representation.

We evaluate VR3D on three aerial-ground person re-identification datasets, including  CARGO \cite{zhang2024view}, AG-ReID.v1 \cite{nguyen2023aerial} and AG-ReID.v2 \cite{nguyen2024ag}. Experimental results demonstrate the promising performance of VR3D compared with recent methods, \emph{e.g.}, it improves Rank-1 and mAP by 5.63\% and 4.93\%, respectively, on CARGO under the A$\leftrightarrow$G protocol.
The main contributions of this paper are summarized as follows:
\begin{itemize}
    \item We propose VR3D, a View-Robust 3D Representation Learning framework, which achieves feature interaction and representation within a unified 3D space to overcome severe aerial-ground viewpoint bias.
    \item We propose VR3I that leverages 3D Geometry-Semantic Attention to establish feature interactions based on actual physical locations. Additionally, we introduce RAF to evaluate the sample-specific reliability of multi-source representations and adaptively aggregate them.
    \item Extensive experiments on AG-ReID.v1, AG-ReID.v2, and CARGO demonstrate the superior performance and effectiveness of the proposed method.
\end{itemize}

\begin{figure*}[t]
\centering
\includegraphics[width=\linewidth]{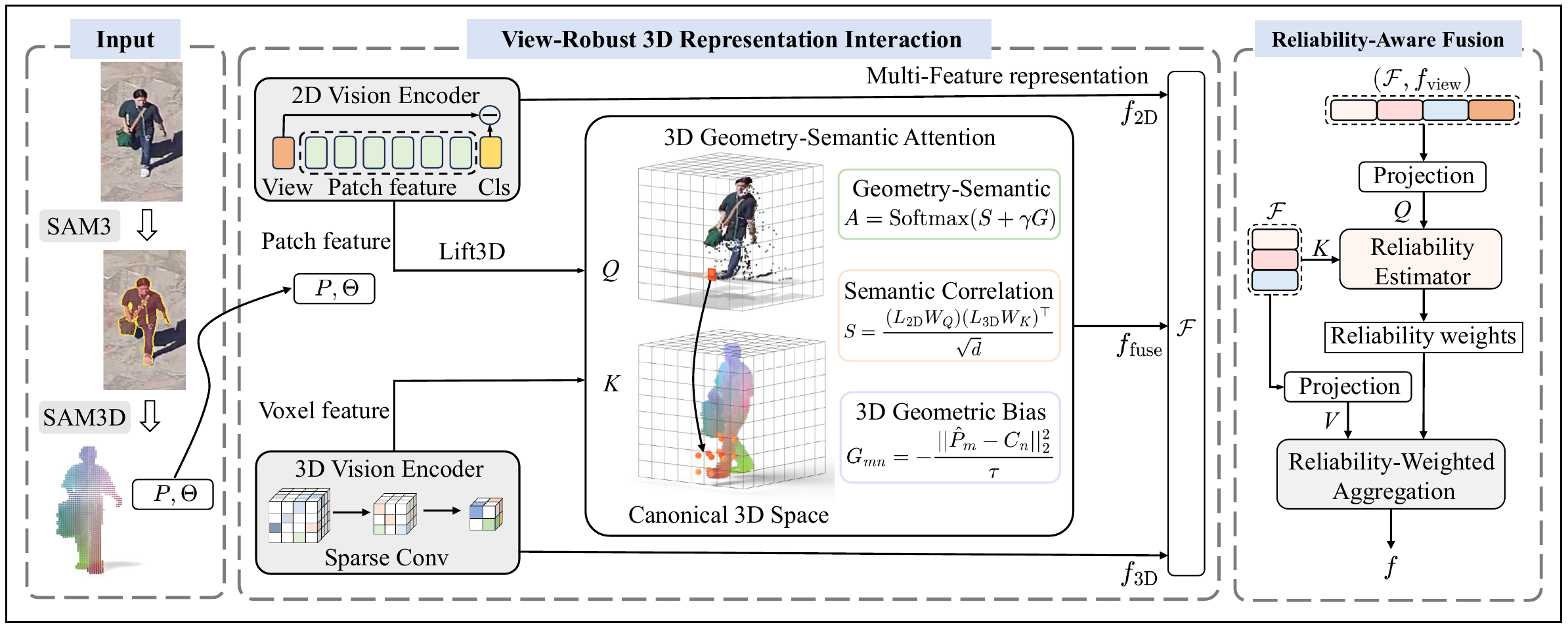} 
\caption{Illustrations of the View-Robust 3D Representation Learning framework (VR3D). An offline pipeline extracts 3D human representation from the input image, followed by the View-Robust 3D Representation Interaction (VR3I) module with 3D Geometry-Semantic Attention for 2D--3D interaction in a 3D canonical space to learn three complementary representations, which are adaptively aggregated by the Reliability-Aware Fusion (RAF) module based on their sample-specific reliability.}
\label{fig:2}
\end{figure*}

\section{Related Work}
\subsection{Aerial-Ground Person Re-Identification}
Aerial-ground person re-identification aims to retrieve the same person across aerial and ground cameras, unlike traditional ReID that mainly focuses on camera networks within a single platform~\cite{cho2022part,lee2023camera,li2025breaking,wang2025idea}. Large cross-platform viewpoint differences cause substantial appearance variations, making this task highly challenging. Existing methods mainly follow two directions.
The first direction learns identity representations through viewpoint-based feature disentanglement. VDT~\cite{zhang2024view}, built upon ViT~\cite{dosovitskiy2021an}, separates view-invariant features from view-related features. Building on this disentangled representation, DTST~\cite{wang2025dynamic} dynamically selects identity-relevant tokens using local and global information, while SeCap~\cite{wang2025secap} introduces adaptive prompts to capture local view-invariant cues. ViSA~\cite{zhang2026view} further models and integrates the disentangled view-invariant and view-specific features to establish cross-view semantic correspondences.
The second direction introduces complementary cross-view information to compensate for incomplete identity cues from a single viewpoint. LATex~\cite{zhang2025latex} and SVPR-ReID~\cite{zheng2026semantic} use CLIP~\cite{radford2021learning} to encode textual information as cross-view semantic guidance. Other methods employ visual generative models. SD-ReID~\cite{wang2026sd} uses Stable Diffusion~\cite{rombach2022high} to obtain complementary view-specific features, while 3D-LENS~\cite{grolleau20263d} employs Hunyuan3D \cite{zhao2025hunyuan3d} for 3D reconstruction and novel-view rendering, generating target-view 2D images for data augmentation.
These methods still alleviate viewpoint variations mainly in the 2D image space. In contrast, we investigate identity feature learning from the perspective of view-robust 3D representations.

\subsection{3D Representation Learning for View Robustness}

Compared with 2D appearance features, 3D representations provide more view-robust information and help alleviate appearance variations caused by viewpoint changes~\cite{shang2022learning}. This advantage has been exploited in several vision tasks. In robotic manipulation, 3D Diffuser Actor~\cite{pmlr-v270-ke25a}, GeoVLA~\cite{sun2025geovla}, and Dexterity-BEV~\cite{zhou2026dexterity} lift 2D observations into 3D representations to improve policy adaptability to camera viewpoint changes. In pose-robust face recognition, 2D-3D Attention and Entropy~\cite{peace20252d} and NPTFace~\cite{ran2026nptface} leverage 3D facial geometry to learn stable identity representations. In cross-view geo-localization, Geo$^{2}$~\cite{zhang2026geo2} maps ground-level and aerial images into a shared 3D-aware space to reduce their representation gap. For conventional single-platform person re-identification, OG-Net~\cite{zheng2024parameter} converts 2D person images into colored 3D point clouds and directly learns identity representations in 3D space, mitigating viewpoint changes and partial occlusion. However, these methods have not explored 3D representations for aerial-ground person re-identification. Moreover, unlike OG-Net, which relies solely on reconstructed point clouds and loses detailed appearance information during 2D-to-3D conversion, our method integrates 2D appearance semantics with 3D human representations in a canonical 3D space to obtain more complete and view-robust 3D representations.

\section{Method}
\subsection{Overview}
Given a query person image $I$ and a gallery image set $\mathcal{G}$ with substantial aerial-ground viewpoint differences from $I$, aerial-ground person re-identification aims to retrieve person images from $\mathcal{G}$ that belong to the same identity as $I$. Existing methods attempt to use a shared encoder to encode images from different viewpoints. However, severe occlusion and geometric deformation introduce strong viewpoint bias into 2D observations, making it exceedingly difficult for the network to learn identity representations. To address this limitation, we propose to map images from different viewpoints into a unified 3D coordinate space to achieve direct feature interaction and representation. Based on this insight, we propose the View-Robust 3D Representation Learning (VR3D) framework, as shown in Fig.~\ref{fig:2}, to learn view-robust identity representations in this unified space.

Obtaining 3D geometric information from a single 2D image is nontrivial. Leveraging extensive 3D knowledge learned from large-scale data, SAM3D can infer 3D representations of occluded and unobserved regions from a 2D observation, thereby complementing the 3D information missing at the current viewpoint. To this end, we construct an offline 3D human representation generation pipeline that takes the original image $I$ and the target person mask $M$ obtained by SAM3 as input and feeds them into SAM3D. The pipeline extracts the voxel-based 3D human representation $V$, the 3D coordinates of the image patches $P$, and the camera pose parameters $\Theta$:
\begin{equation}
\begin{aligned}
M &= \text{SAM3}(I), \\
V, P, \Theta &= \text{SAM3D}(I,M),
\end{aligned}
\end{equation}
where $\Theta=\{R, T, s\}$, with $R \in SO(3)$ and $T \in \mathbb{R}^3$ denoting the rotation matrix and translation vector of the camera extrinsic parameters, respectively, and $s$ is the scale factor.

Although the extracted 3D priors capture geometric information, they lack the appearance semantics required for identity recognition. To align the rich 2D appearance with the 3D geometry, we propose the View-Robust 3D Representation Interaction (VR3I). Utilizing the extracted 3D coordinates and camera pose parameters, VR3I lifts local 2D features into a canonical 3D space shared with the 3D voxels. It then employs 3D Geometry-Semantic Attention to establish interactions between 2D patches and 3D voxels from corresponding body parts based on their physical spatial locations. VR3I leads to three distinct representations: a 2D appearance feature $f_{\text{2D}}$, a spatially-grounded fused feature $f_{\text{fuse}}$, and a view-robust 3D structural feature $f_{\text{3D}}$:
\begin{equation}
\mathcal{F} = \text{VR3I}(I,V,P,\Theta),
\end{equation}
where $\mathcal{F}=\{f_{\text{2D}},f_{\text{fuse}},f_{\text{3D}}\}$.

The reliability of the three representations varies across samples. Extreme viewing angles reduce the reliability of 2D appearance features, while reconstruction errors degrade 3D-related representations. To prevent unreliable information from compromising identity matching, we introduce Reliability-Aware Fusion (RAF), which uses the view-related feature $f_{\text{view}}$ to characterize the viewing condition of each sample and adaptively aggregate the representations in $\mathcal{F}$:
\begin{equation}
f
=
\mathrm{RAF}\left(\mathcal{F},f_{\text{view}}\right).
\end{equation}

\subsection{View-Robust 3D Representation Interaction}
To align 2D appearance with 3D geometry and bridge the cross-view appearance gap, instead of directly concatenating isolated representations, VR3I lifts local 2D appearance features into a canonical 3D space, employing 3D Geometry-Semantic Attention to directly establish feature interactions between 2D patches and 3D voxels from
corresponding human body parts based on their physical spatial locations.

We first encode the 2D image and the 3D human representation separately. For the 2D branch, we use VDT~\cite{zhang2024view} as the 2D image encoder to extract the 2D global appearance feature $f_{\text{2D}}$, local patch features $L_{\text{2D}}$, and the view-related feature $f_{\text{view}}$:
\begin{equation}
f_{\text{2D}}, L_{\text{2D}}, f_{\text{view}} = \text{2DEncoder}(I).
\end{equation}
For the 3D branch, we employ a lightweight sparse 3D encoder to extract the local 3D geometric features $L_{\text{3D}}$ based on the extracted voxels, followed by average pooling to obtain the view-robust 3D structural feature $f_{\text{3D}}$:
\begin{equation}
\begin{aligned}
L_{\text{3D}} &= \text{3DEncoder}(V), \\
f_{\text{3D}} &= \text{AvgPool}(L_{\text{3D}}).
\end{aligned}
\end{equation}

To establish 2D-3D interaction within a unified canonical space, we utilize the patch-level 3D coordinates $P$ and camera pose parameters $\Theta=\{R, T, s\}$ to lift the 2D image patches. The coordinate of the $m$-th image patch is transformed to the canonical 3D space as:
\begin{equation}
\hat{P}_{m} = \frac{(P_{m}-T)R^{\top}}{s},
\end{equation}
where $\hat{P}_{m}$ denotes the transformed coordinate of the $m$-th image patch, aligning the 2D appearance patches with the 3D geometric voxels.

Following this spatial alignment, we introduce 3D Geometry-Semantic Attention (3D-GSA) to establish feature interactions. 3D-GSA jointly models the geometric proximity and semantic correlation between 2D image patches and 3D voxel features. The attention weights are computed as:
\begin{equation}
A = \text{Softmax}(S+\gamma G),
\end{equation}
where $S$ denotes the semantic correlation, $G$ denotes the 3D geometric bias, and $\gamma$ controls the contribution of the geometric information.

The semantic correlation measures the content relevance between the 2D appearance features and the 3D geometric representation:
\begin{equation}
S = \frac{(L_{\text{2D}}W_Q)(L_{\text{3D}}W_K)^{\top}}{\sqrt{d}},
\end{equation}
where $W_Q$ and $W_K$ are learnable projection matrices and $d$ denotes the feature dimension. The geometric bias provides a structural constraint based on the physical distance between each 2D image patch and 3D voxel in the canonical 3D space:
\begin{equation}
G_{mn} = - \frac{||\hat{P}_{m}-C_{n}||_2^{2}}{\tau},
\end{equation}
where $\hat{P}_{m}$ and $C_n$ denote the coordinates of the $m$-th 2D image patch and the $n$-th 3D voxel, respectively, and $\tau$ is a distance scale factor. By jointly considering appearance semantics and spatial proximity, 3D-GSA grounds the 2D semantics within the stable 3D structural framework.

The spatially-grounded features are then computed as:
\begin{equation}
\begin{aligned}
L_{\text{fuse}} &= L_{\text{2D}} + A(L_{\text{3D}}W_V), \\
f_{\text{fuse}} &= \text{AvgPool}(L_{\text{fuse}}),
\end{aligned}
\end{equation}
where $W_V$ is a learnable projection matrix.

Finally, VR3I yields three distinct representations: the 2D appearance feature $f_{\text{2D}}$, the spatially-grounded fused feature $f_{\text{fuse}}$, and the view-robust 3D structural feature $f_{\text{3D}}$. They jointly form the output of VR3I:
\begin{equation}
\mathcal{F} = \{f_{\text{2D}}, f_{\text{fuse}}, f_{\text{3D}}\}.
\end{equation}
These three representations are subsequently fed into RAF.

\subsection{Reliability-Aware Fusion}

The reliability of $\mathcal{F}$ produced by VR3I varies across samples. The reliability of 2D features depends on the viewing condition of the current sample, while 3D-related representations are also affected by reconstruction errors reflected in their feature states. To adaptively aggregate these features, we propose a Reliability-Aware Fusion module, termed RAF, which estimates the relative reliability of each feature source by jointly considering the multi-source representations and the viewing condition of the current sample.

RAF first constructs a view-conditioned routing query by jointly encoding the multi-source identity features $\mathcal{F}$ and the view-related feature $f_{\text{view}}$:
\begin{equation}
q
=
\Phi_q
\left(\operatorname{Concat}\left(\mathcal{F},
f_{\text{view}}
\right)\right),
\label{eq:raf_query}
\end{equation}
where $\Phi_q$ denotes a projection layer.

To adaptively aggregate the representations, RAF formulates the dynamic fusion as cross-attention. Specifically, the multi-source representations in $\mathcal{F}$ are stacked as the keys $K$, while their corresponding expert projections form the values $V$. The attention between $q$ and $K$ yields sample-specific reliability weights, which are applied to $V$ to obtain the final view-robust identity representation $f$:
\begin{equation}
f
=
\operatorname{Softmax}
\left(
\frac{qK^{\top}}{\sqrt{d}}
\right)V.
\end{equation}

\subsection{Training Loss}
We optimize the model using identity supervision and view-related constraints. The identity loss $\mathcal{L}_{\mathrm{id}}$ consists of the cross-entropy loss and the triplet loss, and is applied to both the 2D appearance feature $f_{\text{2D}}$ and the final identity representation $f$. The view loss $\mathcal{L}_{\mathrm{view}}$ consists of a view classification loss and an orthogonality loss that promotes view disentanglement. The overall objective is formulated as
\begin{equation}
\mathcal{L}
=
\mathcal{L}_{\mathrm{id}}[f_{\mathrm{2D}}]
+
\mathcal{L}_{\mathrm{id}}[f]
+
\lambda \mathcal{L}_{\mathrm{view}},
\end{equation}
where $\lambda$ balances the view-related constraint.

\begin{table*}[t]
\centering
\small
\setlength{\tabcolsep}{12px}
\begin{tabular}{lrrrrrrrr}
\toprule
\multirow{2}{*}{Method}
& \multicolumn{2}{c}{A$\leftrightarrow$G}
& \multicolumn{2}{c}{ALL}
& \multicolumn{2}{c}{G$\leftrightarrow$G}
& \multicolumn{2}{c}{A$\leftrightarrow$A} \\
\cmidrule(lr){2-3}
\cmidrule(lr){4-5}
\cmidrule(lr){6-7}
\cmidrule(lr){8-9}
& R1 & mAP
& R1 & mAP
& R1 & mAP
& R1 & mAP \\
\midrule

VDT~\cite{zhang2024view}
& 45.00 & 42.08
& 60.58 & 54.61
& 76.79 & 71.97
& \underline{82.50} & 64.67 \\

DTST~\cite{wang2025dynamic}
& 50.63 & 43.39
& 64.42 & 55.73
& 78.57 & 72.40
& 80.00 & 63.31 \\

VIF~\cite{khalid2025bridging}
& 51.25 & 44.55
& 65.71 & 57.46
& \underline{83.93} & 74.19
& \underline{82.50} & 66.98 \\

SeCap~\cite{wang2025secap}
& 48.75 & 46.37
& 64.72 & 56.89
& 82.54 & 75.24
& \underline{82.50} & 66.90 \\

SD-ReID~\cite{wang2026sd}
& 53.12 & 46.44
& 65.06 & 57.47
& 81.25 & 74.08
& \underline{82.50} & \underline{67.70} \\

ViSA$^{\dagger}$~\cite{zhang2026view}
& \underline{53.75} & \underline{49.64}
& \underline{67.63} & \underline{61.71}
& \underline{83.93} & \textbf{80.01}
& 80.00 & 66.20 \\

\midrule

\rowcolor{gray!15}
VR3D
& \textbf{59.38} & \textbf{54.57}
& \textbf{71.15} & \textbf{64.48}
& \textbf{84.82} & \underline{78.74}
& \textbf{82.50} & \textbf{70.51} \\

\bottomrule
\end{tabular}
\caption{Comparison with existing methods on CARGO. Results marked with $\dagger$ are re-evaluated by us using the released code under the official CARGO protocol; the remaining results are previously reported results. The best performance is shown in \textbf{bold}, and the second-best performance is \underline{underlined}.}
\label{tab:3}
\end{table*}

\section{Experiments}

\subsection{Datasets and Evaluation Metrics}
We conduct experiments on three aerial-ground person re-identification datasets, including one synthetic dataset, CARGO \cite{zhang2024view} and two real-world datasets, AG-ReID.v1 \cite{nguyen2023aerial} and AG-ReID.v2 \cite{nguyen2024ag}. CARGO contains 108,563 person images of 5,000 identities captured by 8 ground cameras and 5 aerial cameras. AG-ReID.v1 contains 21,893 images of 388 identities captured by one aerial camera and one ground camera, with aerial-view images collected at heights ranging from 15 to 45 meters. AG-ReID.v2 extends AG-ReID.v1 in identity scale and viewpoint diversity, containing 100,502 images of 1,615 identities with 15 types of attribute annotations. Its images are collected from UAVs, surveillance cameras, and smart-glass cameras. We follow the official data splits and evaluation protocols of each dataset, and use R1 accuracy and mean Average Precision (mAP) as evaluation metrics.

\subsection{Implementation Details}
Our method is implemented in PyTorch using the FastReID~\cite{he2023fastreid} framework and trained on a single NVIDIA A100 GPU. We use an ImageNet-pretrained Vision Transformer as the 2D backbone. All images are resized to $256 \times 128$ during training and testing. The patch size and stride are set to $16 \times 16$, with an embedding dimension of 768. The batch size is 64, containing 16 identities with 4 images per identity. The model is trained for 120 epochs using SGD. The learning rate follows a warm-up cosine decay schedule, decreasing from $8 \times 10^{-3}$ to $1.6 \times 10^{-6}$. During testing, neither data augmentation nor re-ranking is applied, and the final identity representation is directly used for retrieval.

\begin{table}
\centering

\small
\setlength{\tabcolsep}{4.0pt}
\renewcommand{\arraystretch}{1.08}

\begin{tabular}{lcccc}
\toprule
\multirow{2}{*}{Method}
& \multicolumn{2}{c}{A$\rightarrow$G}
& \multicolumn{2}{c}{G$\rightarrow$A} \\
\cmidrule(lr){2-3}
\cmidrule(lr){4-5}
& R1 & mAP & R1 & mAP \\
\midrule

Explain~\cite{nguyen2023aerial}
& 81.47 & 72.61
& 82.85 & 73.39 \\

VDT~\cite{zhang2024view}
& 82.91 & 74.44
& 86.59 & 78.57 \\

VIF~\cite{khalid2025bridging}
& 83.75 & 75.22
& \underline{87.32} & 79.19 \\

SeCap~\cite{wang2025secap}
& 84.03 & 76.16
& 87.01 & 78.34 \\

SD-ReID~\cite{wang2026sd}
& 85.16 & 75.40
& 85.97 & 77.02 \\

SVPR-ReID~\cite{zheng2026semantic}
& \underline{85.34} & \underline{77.85}
& \underline{87.32} & \underline{80.55} \\

\midrule

\rowcolor{gray!15}
VR3D
& \textbf{87.39} & \textbf{79.95}
& \textbf{89.60} & \textbf{82.87} \\

\bottomrule
\end{tabular}

\caption{Performance comparison on AG-ReID.v1. The best and second-best results are highlighted in \textbf{bold} and \underline{underlined}, respectively.}
\label{tab:1}
\end{table}

\begin{table*}[t]
\centering
\small
\setlength{\tabcolsep}{13px}
\begin{tabular}{lrrrrrrrr}
\toprule
\multirow{2}{*}{Method}
& \multicolumn{2}{c}{A$\rightarrow$C}
& \multicolumn{2}{c}{A$\rightarrow$W}
& \multicolumn{2}{c}{C$\rightarrow$A}
& \multicolumn{2}{c}{W$\rightarrow$A} \\
\cmidrule(lr){2-3}
\cmidrule(lr){4-5}
\cmidrule(lr){6-7}
\cmidrule(lr){8-9}
& R1 & mAP
& R1 & mAP
& R1 & mAP
& R1 & mAP \\
\midrule

Explain~\cite{nguyen2023aerial}
& 87.70 & 79.00
& \textbf{93.67} & 83.14
& 87.35 & 78.24
& 87.73 & 79.08 \\

VDT~\cite{zhang2024view}
& 86.46 & 79.13
& 90.00 & 82.21
& 86.14 & 78.12
& 85.26 & 78.52 \\

V2E~\cite{nguyen2024ag}
& 88.77 & 80.72
& \underline{93.62} & 84.85
& 87.86 & 78.51
& 88.61 & 80.11 \\

SeCap~\cite{wang2025secap}
& 88.12 & 80.84
& 91.44 & 84.01
& 88.24 & 79.99
& 87.56 & 80.15 \\

SD-ReID~\cite{wang2026sd}
& 87.04 & 80.61
& 90.86 & 84.06
& 86.74 & 79.24
& 86.79 & 80.12 \\

ViSA~\cite{zhang2026view}
& \underline{89.43} & \underline{83.61}
& 91.63 & \underline{85.99}
& \underline{88.57} & \underline{82.32}
& \underline{89.23} & \underline{83.10} \\

\midrule

\rowcolor{gray!15}
VR3D
& \textbf{89.98} & \textbf{84.54}
& 93.57 & \textbf{87.96}
& \textbf{90.56} & \textbf{84.39}
& \textbf{89.70} & \textbf{84.17} \\

\bottomrule
\end{tabular}
\caption{Performance comparison on AG-ReID.v2. C, W, and A denote CCTV, wearable, and aerial views, respectively. The best and second-best results are highlighted in \textbf{bold} and \underline{underlined}, respectively.}
\label{tab:2}
\end{table*}

\subsection{Main Results}

Table~\ref{tab:3} reports the results on CARGO. Under the most challenging A$\leftrightarrow$G cross-view setting, our method improves R1 and mAP over the previous state-of-the-art method by 5.63\% and 4.93\%, respectively. This result shows that our method has a clear advantage in aerial-ground cross-view retrieval.

Table~\ref{tab:1} reports the results on AG-ReID.v1. Our method achieves the best performance under both the A$\rightarrow$G and G$\rightarrow$A protocols, reaching R1/mAP of 87.39\%/79.95\% and 89.60\%/82.87\%, respectively. Compared with the previous best method, our method improves both R1 and mAP by more than 2\% under the two protocols. Compared with SD-ReID, which also uses generative priors to complement cross-view information, our method improves mAP by 4.55\% under A$\rightarrow$G and 5.85\% under G$\rightarrow$A. These results show that introducing 3D human representations effectively improves cross-view identity representation learning.

As shown in Table~\ref{tab:2}, our method achieves the best overall performance across the four protocols on AG-ReID.v2. Compared with the previous state-of-the-art method, the largest mAP improvement is 2.07\% under the C$\rightarrow$A protocol. Compared with SD-ReID, our method improves mAP by 3.93\%, 3.90\%, 5.15\%, and 4.05\% under the A$\rightarrow$C, A$\rightarrow$W, C$\rightarrow$A, and W$\rightarrow$A protocols, respectively. These results demonstrate the effectiveness of the proposed method across diverse viewpoints, including aerial cameras, fixed surveillance cameras, and wearable devices.

Overall, our method achieves performance improvements across multiple aerial-ground person re-identification datasets and evaluation protocols. These results demonstrate that our proposed method for learning visual representations in 3D space can learn more viewpoint-robust and discriminative identity features.

\subsection{Ablation Study}

\noindent\textbf{Effects of Key Components.}
To evaluate the effectiveness of each component, we conduct ablation studies in Table~\ref{tab:4}. The 2D baseline uses VDT with only the 2D appearance feature $f_{\text{2D}}$ for retrieval. Directly concatenating the 3D global feature brings only limited gains, indicating that simply introducing 3D information cannot effectively solve the viewpoint-induced appearance bias. In contrast, introducing the fused feature obtained through 2D--3D interaction in the 3D space improves mAP over the baseline by 3.76\% and 2.42\% under the two protocols, respectively. This result shows that our method can learn more robust cross-view identity representations. Based on the fused feature, introducing three complementary feature representations further improves the performance, which verifies their complementarity. Finally, the full model with RAF achieves the best performance and improves R1 by 2.70\% over fixed average fusion under the G$\rightarrow$A protocol, showing that reliability-aware fusion can adaptively aggregate the representations.
\begin{table}[t]
\centering
\setlength{\tabcolsep}{4pt}
\renewcommand{\arraystretch}{1.08}
\small

\begin{tabular}{clcccc}
\toprule
\multirow{2}{*}{}
& \multirow{2}{*}{Configuration}
& \multicolumn{2}{c}{A$\rightarrow$G}
& \multicolumn{2}{c}{G$\rightarrow$A} \\
\cmidrule(lr){3-4}
\cmidrule(lr){5-6}
&
& R1 & mAP
& R1 & mAP \\
\midrule

A & 2D Baseline
& 82.91 & 74.44
& 86.59 & 78.57 \\

B & + 3D Global Feature
& 85.96 & 76.60
& 85.86 & 78.62 \\

C & + VR3I: Fused Feature
& 86.06 & 78.20
& 87.53 & 80.99 \\

D & + VR3I: Multi-Features
& 86.44 & 79.39
& 86.90 & 82.18 \\

\rowcolor{gray!15}
E & + VR3I + RAF
& \textbf{87.39} & \textbf{79.95}
& \textbf{89.60} & \textbf{82.87} \\

\bottomrule
\end{tabular}

\caption{Ablation study of key components on AG-ReID.v1.``3D Global Feature'' directly introduces the global 3D representation $f_{\text{3D}}$, ``Fused Feature'' denotes the fuse representation $f_{\text{fuse}}$ learned by VR3I, and ``Multi-Features'' denotes average fusion of $f_{\text{2D}}$, $f_{\text{fuse}}$, and $f_{\text{3D}}$. RAF further performs reliability-aware fusion.}
\label{tab:4}
\end{table}

\noindent\textbf{Effects of Key Components in 3D-GSA.}
To examine the roles of geometry and semantics in 3D-GSA, we compare different interaction strategies in Table~\ref{tab:5}. Using either geometric bias or semantic correlation alone outperforms coarse-grained global fusion, showing that both spatial localization and semantic correlation benefit 2D--3D interaction. Geometry performs better when used alone, indicating that physical spatial relations provide effective constraints for feature interaction. Combining both cues achieves the best results, confirming their complementarity.

\begin{table}
\centering
\small
\setlength{\tabcolsep}{8.0pt}
\renewcommand{\arraystretch}{1.08}

\begin{tabular}{clcccc}
\toprule
\multirow{2}{*}{}
& \multirow{2}{*}{Interaction}
& \multicolumn{2}{c}{A$\rightarrow$G}
& \multicolumn{2}{c}{G$\rightarrow$A} \\
\cmidrule(lr){3-4}
\cmidrule(lr){5-6}
& & R1 & mAP & R1 & mAP \\
\midrule

A & Coarse
& 86.53 & 78.93
& 86.90 & 81.15 \\

B & Geometry
& 87.11  & 79.60
& 88.46 & 82.15 \\

C & Semantic
& 86.82 & 79.48
& 87.84 & 81.74 \\

\rowcolor{gray!15}
D & Geo. + Sem.
& \textbf{87.39} & \textbf{79.95}
& \textbf{89.60} & \textbf{82.87} \\

\bottomrule
\end{tabular}

\caption{Ablation study of different interaction strategies in 3D-GSA on AG-ReID.v1. ``Coarse'' denotes fusion by directly adding the global 2D and 3D features, while ``Geometry'' and ``Semantic'' use only geometric bias and semantic correlation, respectively. ``Geo. + Sem.'' combines both cues in 3D-GSA.}
\label{tab:5}
\end{table}

\noindent\textbf{Effects of Key Components in RAF.}
RAF is designed to adaptively aggregate representations with different reliability. As shown in Table~\ref{tab:6}, cross-attention outperforms fixed average fusion through adaptive aggregation of the three representations. Further incorporating the view-related feature consistently improves the results, showing that view information provides useful conditions for reliability estimation.

\begin{table}[t]
\centering

\small
\setlength{\tabcolsep}{5pt}
\renewcommand{\arraystretch}{1.08}

\begin{tabular}{cccccccc}
\toprule
\multirow{2}{*}{}
& \multicolumn{3}{c}{Configuration}
& \multicolumn{2}{c}{A$\rightarrow$G}
& \multicolumn{2}{c}{G$\rightarrow$A} \\
\cmidrule(lr){2-4}
\cmidrule(lr){5-6}
\cmidrule(lr){7-8}
& Average & CA & View
& R1 & mAP
& R1 & mAP \\
\midrule

A
& \checkmark &  &
& 86.44 & 79.39
& 86.90 & 82.18 \\

B
&  & \checkmark &
& 87.01 & 79.54
& 88.46 & 82.15 \\

\rowcolor{gray!15}
C
&  & \checkmark & \checkmark
& \textbf{87.39} & \textbf{79.95}
& \textbf{89.60} & \textbf{82.87} \\

\bottomrule
\end{tabular}

\caption{Ablation study of RAF on AG-ReID.v1. ``Average'' uses fixed average fusion, ``CA'' performs adaptive fusion with cross-attention, and ``View'' further conditions the fusion on the view-related feature.}
\label{tab:6}
\end{table}

\begin{figure}[t]
\centering
\includegraphics[width=\linewidth]{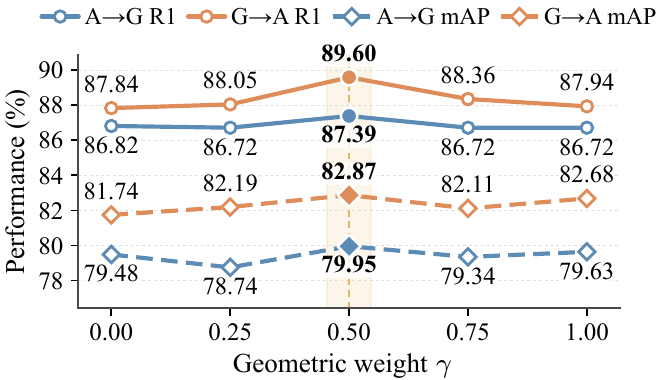}
\caption{Hyperparameter analysis of the geometric weight $\gamma$ on AG-ReID.v1.}
\label{fig:6}
\end{figure}

\noindent\textbf{Hyperparameter Analysis.}
Fig.~\ref{fig:6} presents the hyperparameter analysis of the geometric weight $\gamma$, which controls the physical spatial localization guidance in 3D-GSA. A small $\gamma$ yields insufficient geometric guidance, while a large $\gamma$ overemphasizes spatial proximity at the expense of semantic relevance. Optimal performance is achieved at $\gamma = 0.50$, which is set as the default.

\subsection{Visualization}
\begin{figure}[t]
\centering
\includegraphics[width=\linewidth]{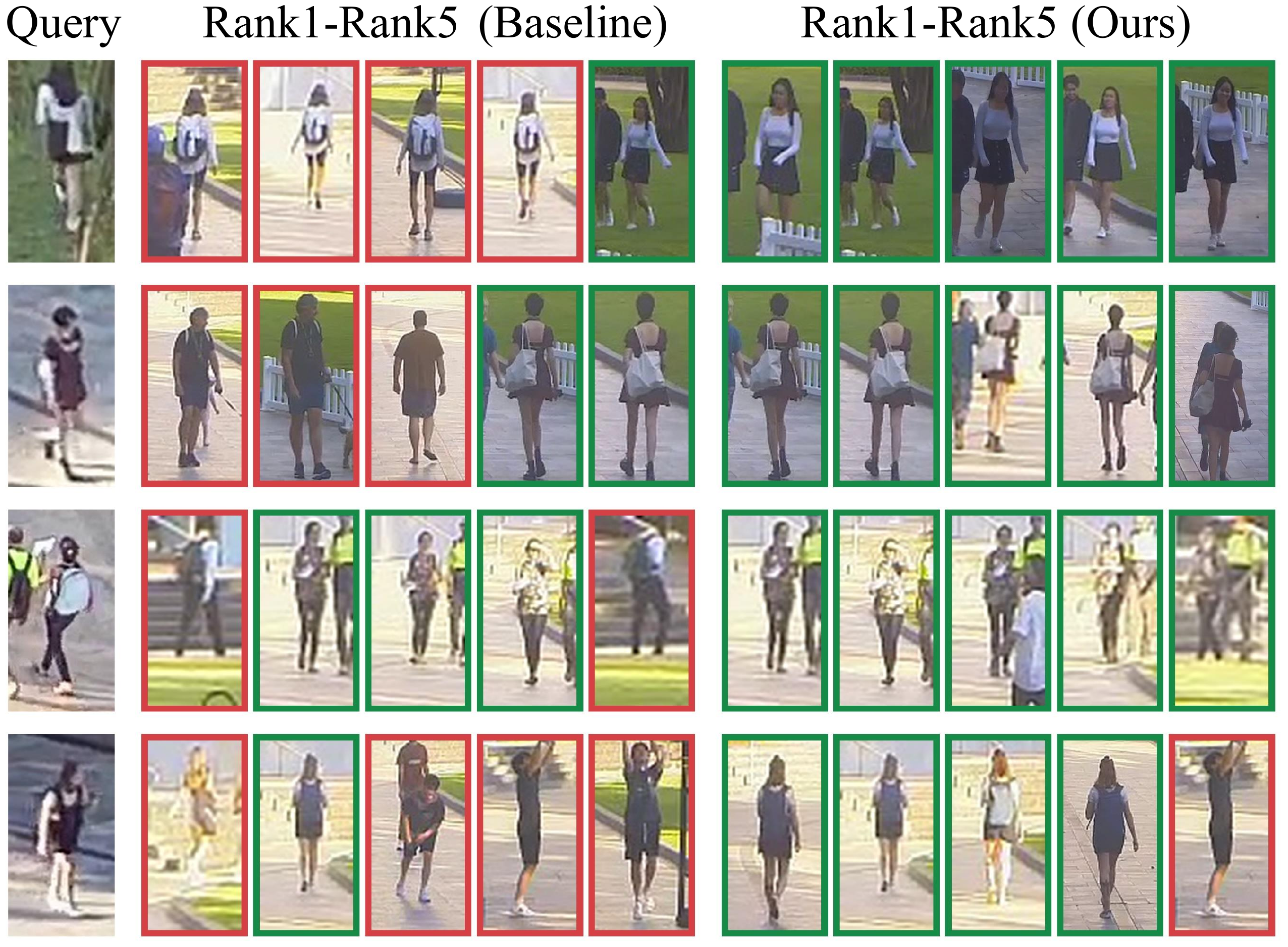}
\caption{Retrieval results on AG-ReID.v1 under the A$\rightarrow$G protocol. Green and red boxes denote correct and incorrect matches, respectively.}
\label{fig:3}
\end{figure}

\noindent\textbf{Rank List Comparison.}
Figure~\ref{fig:3} presents retrieval results on AG-ReID.v1 under the A$\rightarrow$G protocol. Compared with the baseline, our method retrieves more correct cross-view matches among the top-ranked results and reduces interference from visually similar identities, demonstrating more discriminative cross-view identity representations.

\begin{figure}[t]
\centering
\includegraphics[width=\linewidth]{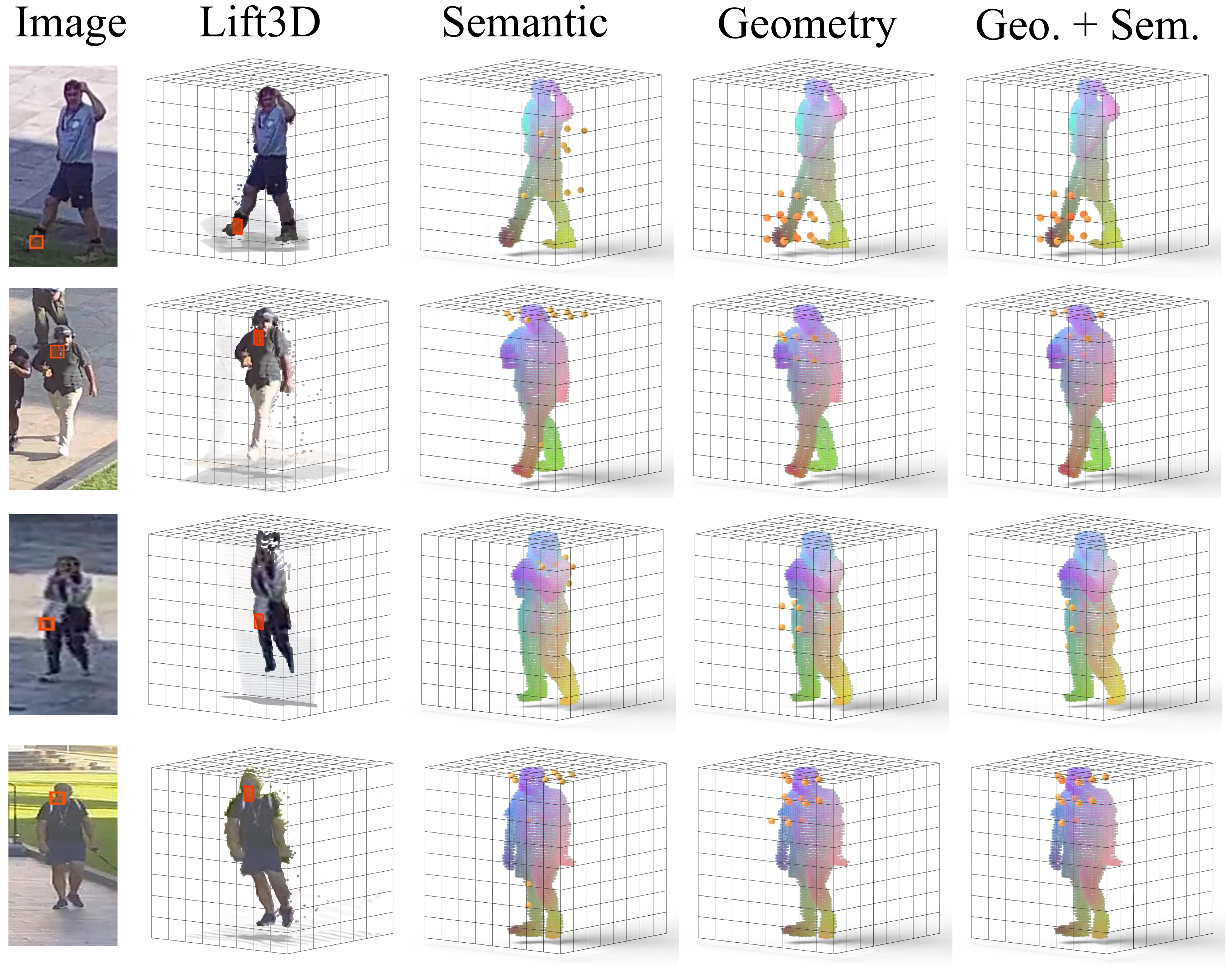}
\caption{Visualization of 3D-GSA on AG-ReID.v1. Orange boxes indicate the query patches, while orange points denote the top-$k$ 3D voxels selected by semantic correlation, geometric bias, and final attention, respectively.}
\label{fig:4}
\end{figure}

\noindent\textbf{3D-GSA Attention Analysis.}
To examine the geometric and semantic cues in 3D-GSA, Fig.~\ref{fig:4} visualizes the top-$k$ voxels selected by semantic correlation, geometric bias, and final attention. Geometric bias favors spatially close voxels but lacks semantic consistency, whereas semantic correlation selects similar regions from unrelated body parts. By combining both cues, the final attention focuses on voxels that are spatially close and semantically consistent with the query patch, enabling reliable 2D--3D interaction in the canonical 3D space.

\begin{figure}[!tbp]
\centering
\includegraphics[width=\linewidth]{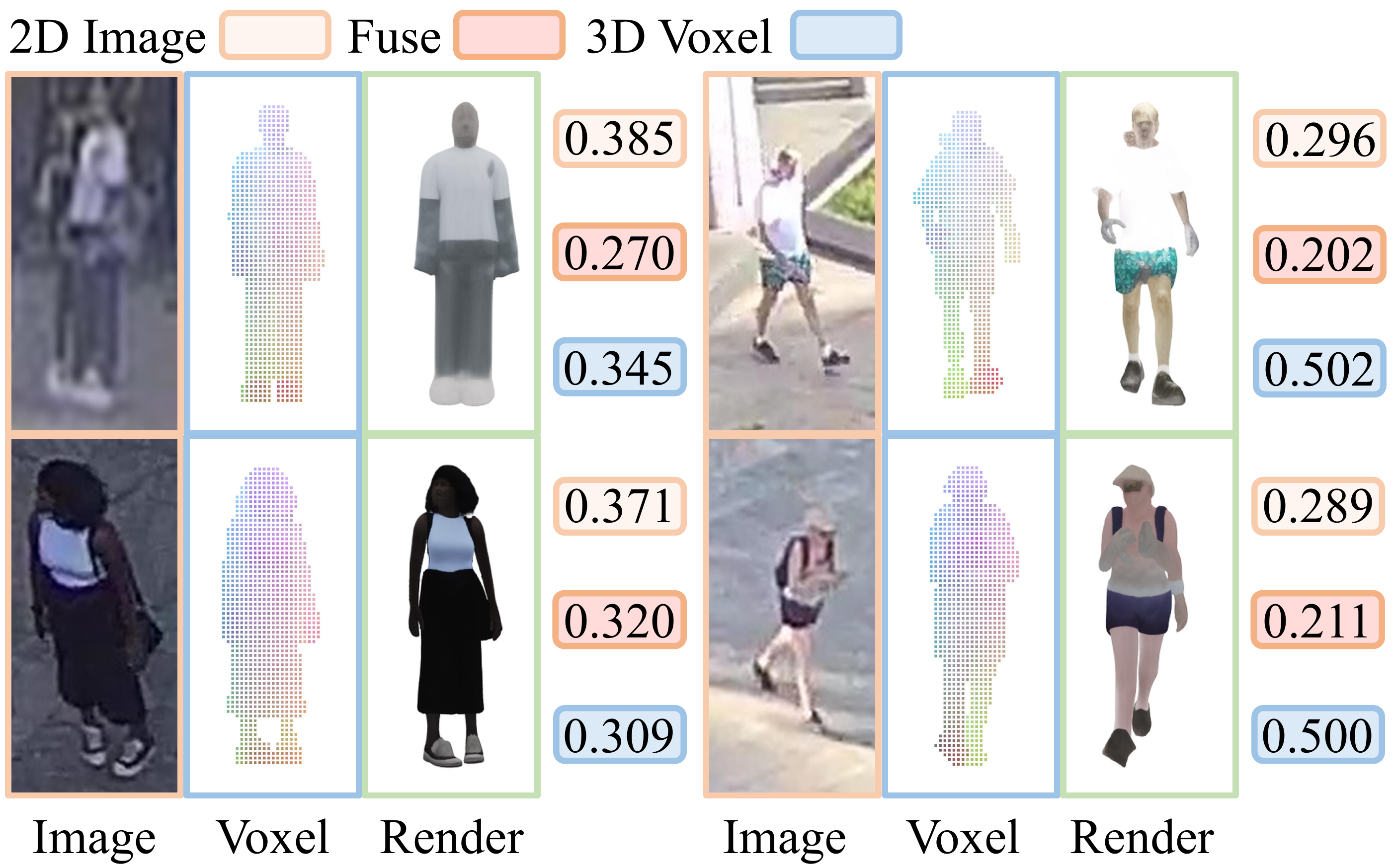}
\caption{Visualization of RAF reliability weights under different samples on AG-ReID.v1.}
\label{fig:5}
\end{figure}

\noindent\textbf{RAF Routing Analysis.}
To verify how RAF estimates the reliability of different feature sources across sample conditions, we visualize its routing weights in Fig.~\ref{fig:5}. RAF gives more weight to $f_{\text{2D}}$ when 3D reconstruction is unreliable, and to $f_{\text{3D}}$ when extreme viewing angles weaken 2D appearance but 3D reconstruction remains reliable. When both 2D appearance and 3D feature are reliable, the contribution of $f_{\text{fuse}}$ increases. These results show that RAF adaptively aggregates the three representations according to their reliability.

\section{Conclusion}

In this work, we propose VR3D, a view-robust 3D representation learning framework for aerial-ground person re-identification. Unlike existing methods that learn representations in a 2D space, VR3D learns identity representations in a view-robust 3D space to reduce the effects of large aerial-ground viewpoint differences. We first construct an offline 3D human representation generation pipeline to obtain 3D information from single-person images. Based on the 3D human representation, VR3I establishes interactions between 2D semantics and 3D human representations in a canonical 3D space, producing multiple complementary identity representations. RAF then adaptively fuses these representations according to their sample-dependent reliability. Experiments on AG-ReID.v1, AG-ReID.v2, and CARGO demonstrate the effectiveness of VR3D.

\bibliography{aaai2027}

@inproceedings{
carion2026sam,
title={{SAM} 3: Segment Anything with Concepts},
author={Nicolas Carion and Laura Gustafson and Yuan-Ting Hu and Shoubhik Debnath and Ronghang Hu and Didac Suris Coll-Vinent and Chaitanya Ryali and Kalyan Vasudev Alwala and Haitham Khedr and Andrew Huang and Jie Lei and Tengyu Ma and Baishan Guo and Arpit Kalla and Markus Marks and Joseph Greer and Meng Wang and Peize Sun and Roman R{\"a}dle and Triantafyllos Afouras and Effrosyni Mavroudi and Katherine Xu and Tsung-Han Wu and Yu Zhou and Liliane Momeni and RISHI HAZRA and Shuangrui Ding and Sagar Vaze and Francois Porcher and Feng Li and Siyuan Li and Aishwarya Kamath and Ho Kei Cheng and Piotr Dollar and Nikhila Ravi and Kate Saenko and Pengchuan Zhang and Christoph Feichtenhofer},
booktitle={The Fourteenth International Conference on Learning Representations},
year={2026},
url={https://openreview.net/forum?id=r35clVtGzw}
}

@InProceedings{Chen_2026_CVPR,
    author    = {Chen, Xingyu and CHU, FU-JEN and Gleize, Pierre and Liang, Kevin J and Sax, Alexander and Tang, Hao and Wang, Weiyao and Guo, Michelle and Hardin, Thibaut and Li, Xiang and Lin, Aohan and Liu, Jia-Wei and Ma, Ziqi and Sagar, Anushka and Song, Bowen and Wang, Xiaodong and Yang, Jianing and Zhang, Bowen and Doll\'ar, Piotr and Gkioxari, Georgia and Feiszli, Matt and Malik, Jitendra},
    title     = {SAM 3D: 3Dfy Anything in Images},
    booktitle = {Proceedings of the IEEE/CVF Conference on Computer Vision and Pattern Recognition (CVPR)},
    month     = {June},
    year      = {2026},
    pages     = {7220-7232}
}

@inproceedings{nguyen2023aerial,
  title={Aerial-ground person re-id},
  author={Nguyen, Huy and Nguyen, Kien and Sridharan, Sridha and Fookes, Clinton},
  booktitle={2023 IEEE International Conference on Multimedia and Expo (ICME)},
  pages={2585--2590},
  year={2023},
  organization={IEEE}
}

@article{nguyen2024ag,
  title={AG-ReID. v2: Bridging aerial and ground views for person re-identification},
  author={Nguyen, Huy and Nguyen, Kien and Sridharan, Sridha and Fookes, Clinton},
  journal={IEEE Transactions on Information Forensics and Security},
  volume={19},
  pages={2896--2908},
  year={2024},
  publisher={IEEE}
}

@inproceedings{zhang2024view,
  title={View-decoupled transformer for person re-identification under aerial-ground camera network},
  author={Zhang, Quan and Wang, Lei and Patel, Vishal M and Xie, Xiaohua and Lai, Jianhaung},
  booktitle={Proceedings of the IEEE/CVF Conference on Computer Vision and Pattern Recognition},
  pages={22000--22009},
  year={2024}
}

@inproceedings{wang2025dynamic,
  title={Dynamic Token Selective Transformer for Aerial-Ground Person Re-Identification},
  author={Wang, Yuhai and Pishgar, Maryam},
  booktitle={2025 IEEE International Conference on Multimedia and Expo (ICME)},
  pages={1--6},
  year={2025},
  organization={IEEE}
}

@inproceedings{wang2025secap,
  title={SeCap: self-calibrating and adaptive prompts for cross-view person re-identification in aerial-ground networks},
  author={Wang, Shining and Wang, Yunlong and Wu, Ruiqi and Jiao, Bingliang and Wang, Wenxuan and Wang, Peng},
  booktitle={Proceedings of the IEEE/CVF Conference on Computer Vision and Pattern Recognition},
  pages={22119--22128},
  year={2025}
}

@inproceedings{khalid2025bridging,
  title={Bridging the sky and ground: Towards view-invariant feature learning for aerial-ground person re-identification},
  author={Khalid, Wajahat and Liu, Bin and Li, Xulin and Waqas, Muhammad and Afgan, Muhammad Sher},
  booktitle={Proceedings of the IEEE/CVF International Conference on Computer Vision},
  pages={9749--9758},
  year={2025}
}

@inproceedings{zhang2026view,
  title={View-Aware Semantic Alignment for Aerial-Ground Person Re-Identification},
  author={Zhang, Quan and Cai, Zeqiang and Zhao, Peiming and Wu, Jingze and Wu, Cailun and Chen, Hongbo and Lai, Jianhuang},
  booktitle={Proceedings of the IEEE/CVF Conference on Computer Vision and Pattern Recognition},
  pages={4383--4392},
  year={2026}
}

@inproceedings{zheng2026semantic,
  title={Semantic-Driven Visual Progressive Refinement for Aerial-Ground Person ReID: A Challenging Large-Scale Benchmark},
  author={Zheng, Aihua and Xie, Hao and Wan, Xixi and Wang, Zi and Li, Shihao and Tang, Jin and Luo, Bin},
  booktitle={Proceedings of the AAAI Conference on Artificial Intelligence},
  volume={40},
  pages={13360--13368},
  year={2026}
}

@article{wang2026sd,
  title={SD-ReID: View-aware Stable Diffusion for Aerial-Ground Person Re-Identification},
  author={Wang, Yuhao and Hu, Xiang and Wang, Lixin and Zhang, Pingping and Lu, Huchuan},
  journal={IEEE Transactions on Image Processing},
  year={2026},
  publisher={IEEE}
}

@article{grolleau20263d,
  title={3D-LENS: A 3D Lifting-based Elevated Novel-view Synthesis method for Single-View Aerial-Ground Re-Identification},
  author={Grolleau, William and Sabourin, Astrid and Lapouge, Guillaume and Achard, Catherine},
  journal={arXiv preprint arXiv:2604.26520},
  year={2026}
}

@article{zheng2024parameter,
  title={Parameter-Efficient Person Re-Identification in the 3D Space},
  author={Zheng, Zhedong and Wang, Xiaohan and Zheng, Nenggan and Yang, Yi},
  journal={IEEE transactions on neural networks and learning systems},
  volume={35},
  number={6},
  pages={7534--7547},
  year={2024}
}

@inproceedings{he2021transreid,
  title={Transreid: Transformer-based object re-identification},
  author={He, Shuting and Luo, Hao and Wang, Pichao and Wang, Fan and Li, Hao and Jiang, Wei},
  booktitle={Proceedings of the IEEE/CVF international conference on computer vision},
  pages={15013--15022},
  year={2021}
}

@inproceedings{li2023clip,
  title={Clip-reid: exploiting vision-language model for image re-identification without concrete text labels},
  author={Li, Siyuan and Sun, Li and Li, Qingli},
  booktitle={Proceedings of the AAAI conference on artificial intelligence},
  volume={37},
  pages={1405--1413},
  year={2023}
}

@inproceedings{he2024instruct,
  title={Instruct-reid: A multi-purpose person re-identification task with instructions},
  author={He, Weizhen and Deng, Yiheng and Tang, Shixiang and Chen, Qihao and Xie, Qingsong and Wang, Yizhou and Bai, Lei and Zhu, Feng and Zhao, Rui and Ouyang, Wanli and others},
  booktitle={Proceedings of the IEEE/CVF Conference on Computer Vision and Pattern Recognition},
  pages={17521--17531},
  year={2024}
}

@inproceedings{yuan2025poses,
  title={From poses to identity: Training-free person re-identification via feature centralization},
  author={Yuan, Chao and Zhang, Guiwei and Ma, Changxiao and Zhang, Tianyi and Niu, Guanglin},
  booktitle={Proceedings of the Computer Vision and Pattern Recognition Conference},
  pages={24409--24418},
  year={2025}
}

@inproceedings{zhou2026hierarchical,
  title={Hierarchical prompt learning for image-and text-based person re-identification},
  author={Zhou, Linhan and Li, Shuang and Dong, Neng and Tai, Yonghang and Zhang, Yafei and Li, Huafeng},
  booktitle={Proceedings of the AAAI Conference on Artificial Intelligence},
  volume={40},
  pages={13728--13736},
  year={2026}
}

@inproceedings{cho2022part,
  title={Part-based pseudo label refinement for unsupervised person re-identification},
  author={Cho, Yoonki and Kim, Woo Jae and Hong, Seunghoon and Yoon, Sung-Eui},
  booktitle={Proceedings of the IEEE/CVF conference on computer vision and pattern recognition},
  pages={7308--7318},
  year={2022}
}

@inproceedings{lee2023camera,
  title={Camera-driven representation learning for unsupervised domain adaptive person re-identification},
  author={Lee, Geon and Lee, Sanghoon and Kim, Dohyung and Shin, Younghoon and Yoon, Yongsang and Ham, Bumsub},
  booktitle={Proceedings of the IEEE/CVF International Conference on Computer Vision},
  pages={11453--11462},
  year={2023}
}

@article{li2025breaking,
  title={Breaking the paired sample barrier in person re-identification: Leveraging unpaired samples for domain generalization},
  author={Li, Huafeng and Liu, Yaoxin and Zhang, Yafei and Li, Jinxing and Yu, Zhengtao},
  journal={IEEE Transactions on Information Forensics and Security},
  year={2025},
  publisher={IEEE}
}

@inproceedings{wang2025idea,
  title={Idea: Inverted text with cooperative deformable aggregation for multi-modal object re-identification},
  author={Wang, Yuhao and Lv, Yongfeng and Zhang, Pingping and Lu, Huchuan},
  booktitle={Proceedings of the Computer Vision and Pattern Recognition Conference},
  pages={29701--29710},
  year={2025}
}

@inproceedings{zhang2023ground,
  title={Ground-to-aerial person search: Benchmark dataset and approach},
  author={Zhang, Shizhou and Yang, Qingchun and Cheng, De and Xing, Yinghui and Liang, Guoqiang and Wang, Peng and Zhang, Yanning},
  booktitle={Proceedings of the 31st ACM International Conference on Multimedia},
  pages={789--799},
  year={2023}
}

@inproceedings{nguyen2025ag,
  title={Ag-vpreid: A challenging large-scale benchmark for aerial-ground video-based person re-identification},
  author={Nguyen, Huy and Nguyen, Kien and Pemasiri, Akila and Liu, Feng and Sridharan, Sridha and Fookes, Clinton},
  booktitle={Proceedings of the Computer Vision and Pattern Recognition Conference},
  pages={1241--1251},
  year={2025}
}

@article{zhang2025latex,
  title={Latex: Leveraging attribute-based text knowledge for aerial-ground person re-identification},
  author={Zhang, Pingping and Hu, Xiang and Wang, Yuhao and Lu, Huchuan},
  journal={arXiv preprint arXiv:2503.23722},
  year={2025}
}

@article{shang2022learning,
  title={Learning viewpoint-agnostic visual representations by recovering tokens in 3d space},
  author={Shang, Jinghuan and Das, Srijan and Ryoo, Michael},
  journal={Advances in Neural Information Processing Systems},
  volume={35},
  pages={31031--31044},
  year={2022}
}

@inproceedings{
dosovitskiy2021an,
title={An Image is Worth 16x16 Words: Transformers for Image Recognition at Scale},
author={Alexey Dosovitskiy and Lucas Beyer and Alexander Kolesnikov and Dirk Weissenborn and Xiaohua Zhai and Thomas Unterthiner and Mostafa Dehghani and Matthias Minderer and Georg Heigold and Sylvain Gelly and Jakob Uszkoreit and Neil Houlsby},
booktitle={International Conference on Learning Representations},
year={2021},
url={https://openreview.net/forum?id=YicbFdNTTy}
}

@inproceedings{radford2021learning,
  title={Learning transferable visual models from natural language supervision},
  author={Radford, Alec and Kim, Jong Wook and Hallacy, Chris and Ramesh, Aditya and Goh, Gabriel and Agarwal, Sandhini and Sastry, Girish and Askell, Amanda and Mishkin, Pamela and Clark, Jack and others},
  booktitle={International conference on machine learning},
  pages={8748--8763},
  year={2021},
  organization={PmLR}
}

@article{zhao2025hunyuan3d,
  title={Hunyuan3d 2.0: Scaling diffusion models for high resolution textured 3d assets generation},
  author={Zhao, Zibo and Lai, Zeqiang and Lin, Qingxiang and Zhao, Yunfei and Liu, Haolin and Yang, Shuhui and Feng, Yifei and Yang, Mingxin and Zhang, Sheng and Yang, Xianghui and others},
  journal={arXiv preprint arXiv:2501.12202},
  year={2025}
}

@inproceedings{rombach2022high,
  title={High-resolution image synthesis with latent diffusion models},
  author={Rombach, Robin and Blattmann, Andreas and Lorenz, Dominik and Esser, Patrick and Ommer, Bj{\"o}rn},
  booktitle={Proceedings of the IEEE/CVF conference on computer vision and pattern recognition},
  pages={10684--10695},
  year={2022}
}

@article{sun2025geovla,
  title={Geovla: Empowering 3d representations in vision-language-action models},
  author={Sun, Lin and Xie, Bin and Liu, Yingfei and Shi, Hao and Wang, Tiancai and Cao, Jiale},
  journal={arXiv preprint arXiv:2508.09071},
  year={2025}
}

@article{zhou2026dexterity,
  title={Dexterity-BEV: Aligning 3D World and Actions for Generalizable Robot Policies Learning},
  author={Zhou, Huayi and Gao, Wei and Lu, Dekun and Liu, Ruiji and Zhang, Zhanqi and Zhang, Ziyang and Chen, Jian and Zhou, Wenlve and Xu, Sheng and Li, Shumin and others},
  journal={arXiv preprint arXiv:2606.02274},
  year={2026}
}

@InProceedings{pmlr-v270-ke25a,
  title = 	 {3D Diffuser Actor: Policy Diffusion with 3D Scene Representations},
  author =       {Ke, Tsung-Wei and Gkanatsios, Nikolaos and Fragkiadaki, Katerina},
  booktitle = 	 {Proceedings of The 8th Conference on Robot Learning},
  pages = 	 {1949--1974},
  year = 	 {2025},
  editor = 	 {Agrawal, Pulkit and Kroemer, Oliver and Burgard, Wolfram},
  volume = 	 {270},
  series = 	 {Proceedings of Machine Learning Research},
  month = 	 {06--09 Nov},
  publisher =    {PMLR}
}

@inproceedings{ran2026nptface,
  title={NPTFace: Native Pose-aligned Transformer for Face Recognition},
  author={Ran, Zimin and Zhu, Xuhan and An, Xiang and Ren, Xingyu and Yang, Kaicheng and Tang, Feilong and Chen, Zhichao and Wang, Yumeng and Feng, Ziyong and Wang, Xianzhi and others},
  booktitle={Proceedings of the IEEE/CVF Conference on Computer Vision and Pattern Recognition},
  pages={1204--1213},
  year={2026}
}

@inproceedings{peace20252d,
  title={2D-3D Attention and Entropy for Pose Robust 2D Facial Recognition},
  author={Peace, J Brennan and Hu, Shuowen and Riggan, Benjamin S},
  booktitle={2025 IEEE 19th International Conference on Automatic Face and Gesture Recognition (FG)},
  pages={1--11},
  year={2025},
  organization={IEEE}
}

@inproceedings{zhang2026geo2,
  title={Geo2: Geometry-Guided Cross-view Geo-Localization and Image Synthesis},
  author={Zhang, Yancheng and Zhang, Xiaohan and Sun, Guangyu and Lyu, Zonglin and Wshah, Safwan and Chen, Chen},
  booktitle={Proceedings of the IEEE/CVF Conference on Computer Vision and Pattern Recognition},
  pages={19432--19442},
  year={2026}
}

@inproceedings{he2023fastreid,
  title={Fastreid: A pytorch toolbox for general instance re-identification},
  author={He, Lingxiao and Liao, Xingyu and Liu, Wu and Liu, Xinchen and Cheng, Peng and Mei, Tao},
  booktitle={Proceedings of the 31st ACM international conference on multimedia},
  pages={9664--9667},
  year={2023}
}
\clearpage
\appendix

\twocolumn[
\begin{center}
    {\LARGE\bfseries Supplementary Material}
\end{center}
\vspace{1em}
]

\section{Introduction}

In this supplementary material, we provide additional implementation details and experimental results to complement the main paper. Specifically, the supplementary material is organized as follows:

\begin{enumerate}
    \item \textbf{Implementation details:}
    \begin{itemize}
        \item Offline 3D human representation generation pipeline
        \item Network details of the sparse 3D encoder, 3D-GSA, and RAF
    \end{itemize}

    \item \textbf{Additional results and visualizations:}
    \begin{itemize}
        \item Additional retrieval results on CARGO and AG-ReID.v2
        \item Additional visualizations of 3D-GSA on CARGO and AG-ReID.v2
        \item Additional visualizations of RAF on CARGO and AG-ReID.v2
    \end{itemize}
\end{enumerate}

\section{Implementation Details}

\subsection{3D Human Representation Generation}

To obtain complete 3D human representations, we construct an offline 3D human representation generation pipeline consisting of two stages: target person mask generation and 3D representation reconstruction. First, SAM3 generates candidate person masks using the \texttt{person} prompt, and the target mask is selected according to its position and area. We further use the \texttt{bag} prompt to incorporate identity-related belongings into the mask, resulting in a more complete target mask.

The original image and the target mask are jointly fed into SAM3D to obtain the voxel-based 3D implicit representation, patch-level 3D coordinates, and pose parameters. To reduce computational and storage costs, we directly retain the implicit features from the decoding process as the 3D human representation.

\begin{figure*}[t]
\centering
\includegraphics[width=0.9\linewidth]{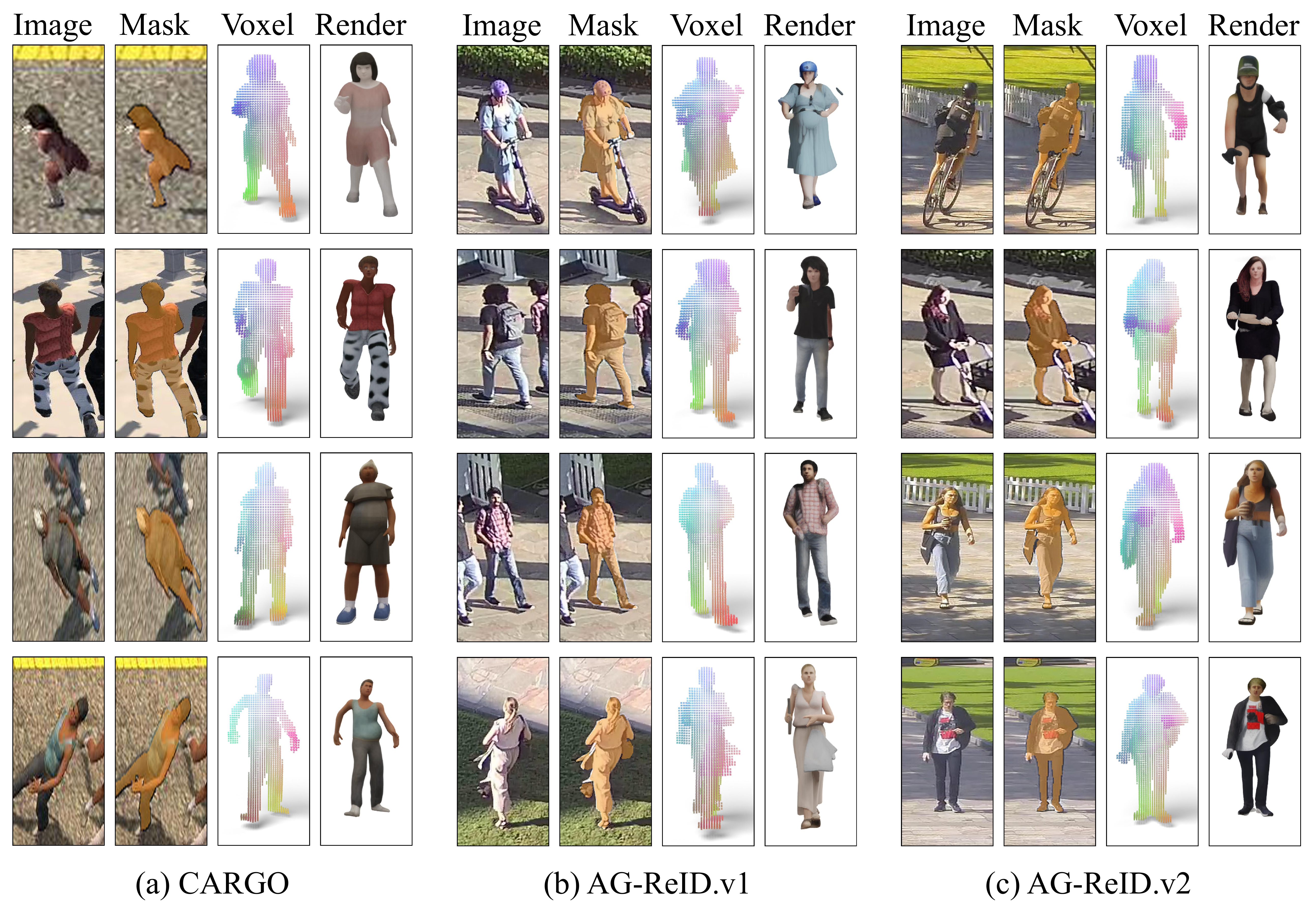}
\caption{Representative 3D person representation generation results on CARGO, AG-ReID.v1, and AG-ReID.v2. From left to right: input image, target person mask, voxel-based 3D implicit representation, and rendered 3D human.}
\label{fig:sup2}
\end{figure*}

As shown in Fig.~\ref{fig:sup2}, SAM3 preserves the target person and identity-related regions, while SAM3D produces complete and consistent 3D visual representations across CARGO, AG-ReID.v1, and AG-ReID.v2, providing structural cues for subsequent 2D--3D interaction. The rendered results are included only for visual inspection of the reconstruction quality.

\subsection{Network Details}

\noindent\textbf{Sparse 3D Encoder.}
Each sample's 3D implicit representation is an 8-channel sparse voxel feature defined on a \(64^3\) lattice. The sparse 3D encoder contains 11 convolutional layers and progressively transforms the input into a 256-channel \(8^3\) sparse voxel representation. Global average pooling over the active voxels produces the 256-dimensional global 3D representation \(f_{\text{3D}}\).

\noindent\textbf{3D-GSA Settings.}
In 3D-GSA, the 768-dimensional 2D patch features form the queries, while the 256-dimensional voxel features are projected to 768 dimensions to form the keys and values. The semantic weight is set to 1.0, the geometric weight to \(\gamma=0.5\), and the distance temperature to \(\tau=0.03\).

\noindent\textbf{RAF Settings.}
RAF uses four attention heads, each with a dimension of \(768/4=192\). The three source representations are first projected into a common 768-dimensional space. The projected representations are concatenated with the 768-dimensional view-related feature \(f_{\text{view}}\) and further projected to form the routing query. They are also used to construct the routing keys and expert features. The reliability-weighted outputs are concatenated and projected to obtain the final identity representation.

\section{Additional Results and Visualizations}

CARGO is a large-scale benchmark collected by multiple aerial and ground cameras, with substantial cross-view variations. AG-ReID.v2 covers UAV, CCTV, and wearable-camera views, providing a complementary evaluation across different camera platforms. To verify the effectiveness of VR3D across these datasets, we visualize its retrieval results, 3D-GSA attention, and RAF routing weights.

\subsection{Retrieval Results}

\begin{figure*}[t]
\centering
\includegraphics[width=\linewidth]{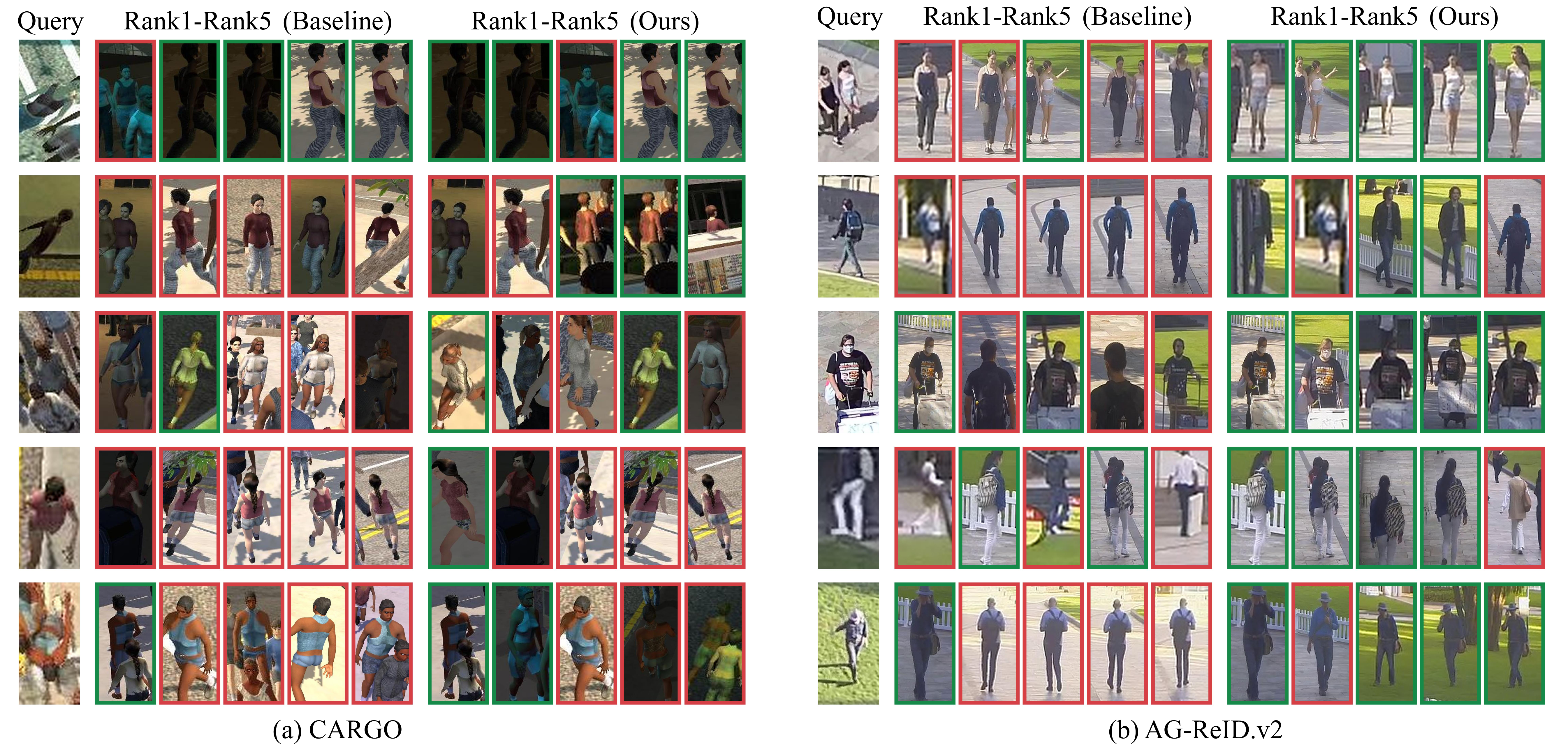}
\caption{Retrieval results of the baseline and VR3D on CARGO under the A$\rightarrow$G protocol and on AG-ReID.v2 under the A$\rightarrow$C protocol.}
\label{fig:app_retrieval}
\end{figure*}

Fig.~\ref{fig:app_retrieval} compares the rank lists of the baseline and VR3D on CARGO under the A$\rightarrow$G protocol and on AG-ReID.v2 under the A$\rightarrow$C protocol. This comparison examines whether the learned 3D representations improve identity matching under large viewpoint changes. VR3D ranks more correct matches ahead of visually similar distractors on both datasets, showing the effectiveness of view-robust 3D representation learning for cross-view retrieval.

\subsection{3D-GSA Visualizations}

\begin{figure*}[t]
\centering
\includegraphics[width=\linewidth]{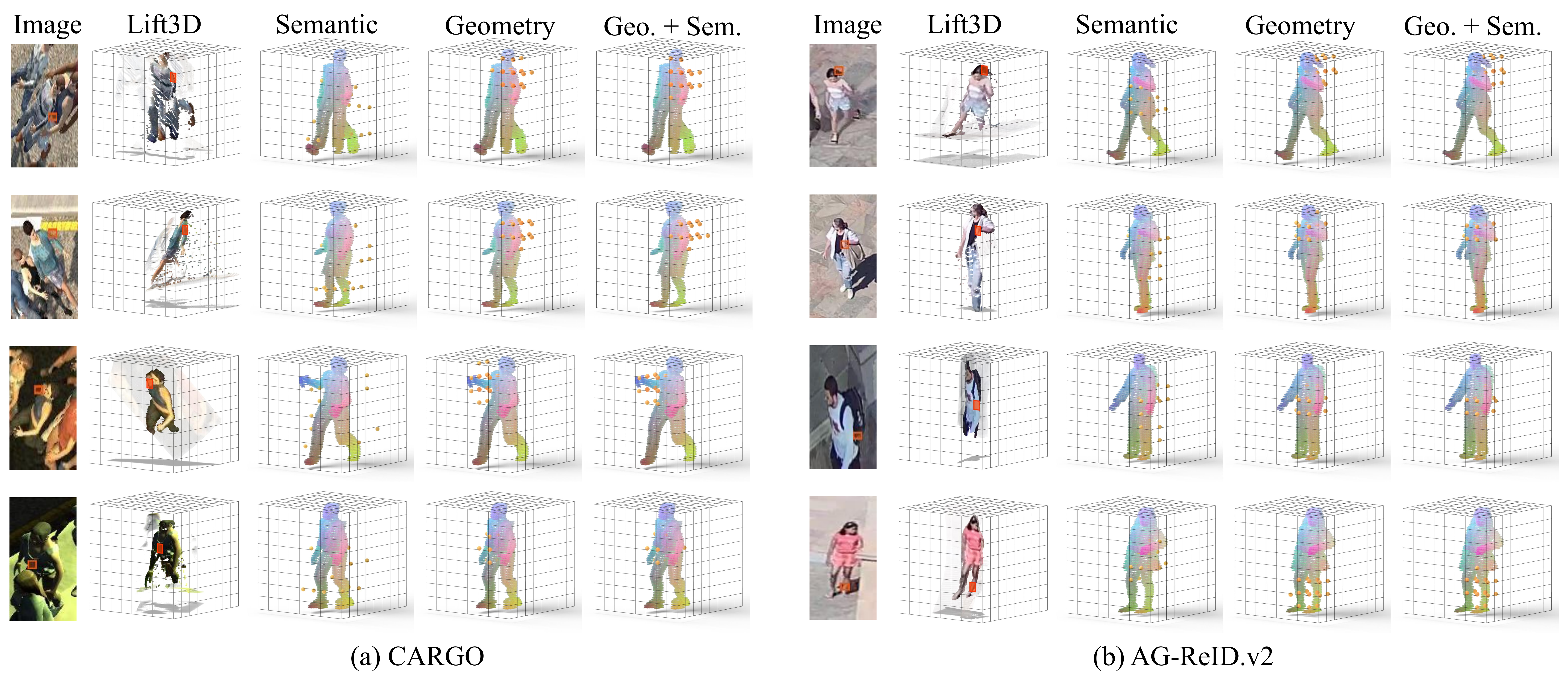}
\caption{Visualization of 3D-GSA on CARGO and AG-ReID.v2. Orange boxes indicate the query patches, while orange points denote the top-$k$ 3D voxels selected by semantic correlation, geometric bias, and final attention, respectively.}
\label{fig:app_3dgsa}
\end{figure*}

The joint use of geometric and semantic cues is important for establishing reliable 2D--3D correspondence. We therefore visualize the geometric bias, semantic correlation, and final attention in Fig.~\ref{fig:app_3dgsa}. Geometric bias favors nearby voxels, while semantic correlation identifies related but sometimes spatially inconsistent regions. The final attention concentrates on voxels supported by both cues, verifying that 3D-GSA performs spatially grounded 2D--3D interaction.

\subsection{RAF Visualizations}

To verify that RAF performs adaptive rather than fixed fusion, Fig.~10 presents the routing weights of the three representations under different sample conditions. In the first CARGO example, the inaccurate 3D reconstruction leads to a low weight for $f_{\mathrm{3D}}$, while $f_{\mathrm{2D}}$ and $f_{\mathrm{fuse}}$ receive higher weights. For samples with reliable 3D structure, RAF increases the contribution of $f_{\mathrm{3D}}$ or assigns more balanced weights. The AG-ReID.v2 examples further show that RAF adjusts the dominant feature source according to the available appearance and structural cues. In the second example, $f_{\mathrm{fuse}}$ receives the highest weight when both 2D and 3D information are useful.

\begin{figure*}[t]
\centering
\includegraphics[width=\linewidth]{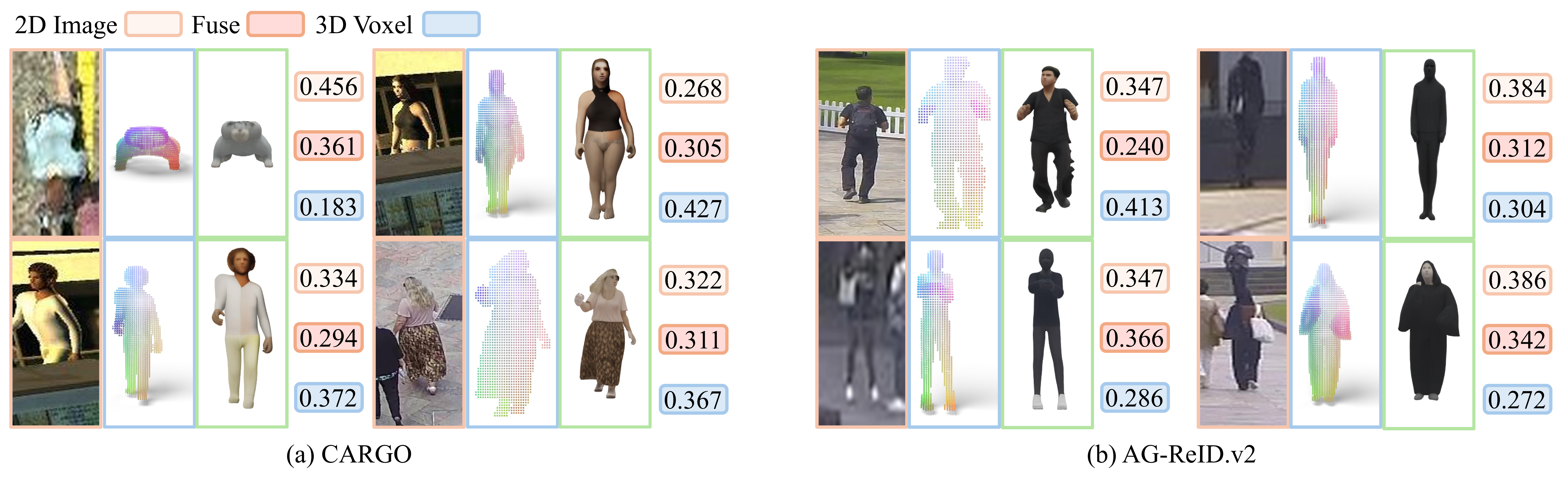}
\caption{Routing weights of RAF on CARGO and AG-ReID.v2.}
\label{fig:app_raf}
\end{figure*}
\end{document}